\documentclass[12pt]{iopart}

\expandafter\let\csname equation*\endcsname=\relax
\expandafter\let\csname endequation*\endcsname=\relax
\usepackage{amsmath,amssymb,amsthm}
\usepackage{soul}\setuldepth{article}
\usepackage{graphicx}
\usepackage{xcolor}
\usepackage{url}

\begin{document}

% \pagestyle{headings}
% \def\thepage{}
% \begin{frontmatter}              % The preamble begins here.

%\pretitle{Pretitle}
\title[]{Respiratory Differencing: Enhancing Pulmonary Thermal Ablation Evaluation Through \\ Pre- and Intra-Operative Image Fusion}

% \markboth{}{November 2024\hb}
% %\subtitle{Subtitle}
\author{Wan~Li$^{1, *}$, Wei~Li$^{2, *}$, Moheng~Rong$^1$, Yutao~Rao$^3$, Hui~Tang$^3$, Yudong~Zhang$^{3, \dag}$, Feng~Wang$^{1, \dag}$}
\address{$^1$ Department of Respiratory and Critical Care Medicine, Beijing Institute of Respiratory Medicine and Beijing Chao Yang Hospital, Capital Medical University, Beijing, CN}
\address{$^2$ First Imaging Medical Equipment, Shanghai, CN}
\address{$^3$ School of Computer Science and Engineering, Southeast University, Nanjing, CN}
\address{$^*$ These authors contributed equally to this work.}
\address{$^\dag$ Co-Corresponding Authors}

\eads{\mailto{yudongzhang@seu.edu.cn}, \mailto{fengwangmail@163.com}}

\vspace{10pt}
\begin{abstract}

CT image-guided thermal ablation is widely used for lung cancer treatment; however, follow-up data indicate that physicians' subjective assessments of intraoperative images often overestimate the ablation effect, potentially leading to incomplete treatment. 

%This overestimation primarily results from respiratory-induced lung deformations and ablation-related image alterations, which compel physicians to rely on spatial intuition to determine whether the treatment area adequately encompasses the preoperatively defined target.

To address these challenges, we developed \textit{Respiratory Differencing}, a novel intraoperative CT image assistance system aimed at improving ablation evaluation. The system first segments tumor regions in preoperative CT images and then employs a multi-stage registration process to align these images with corresponding intraoperative or postoperative images, compensating for respiratory deformations and treatment-induced changes. This system provides two key outputs to help physicians evaluate intraoperative ablation. First, differential images are generated by subtracting the registered preoperative images from the intraoperative ones, allowing direct visualization and quantitative comparison of pre- and post-treatment differences. These differential images enable physicians to assess the relative positions of the tumor and ablation zones, even when the tumor is no longer visible in post-ablation images, thus improving the subjective evaluation of ablation effectiveness. Second, the system provides a quantitative metric that measures the discrepancies between the tumor area and the treatment zone, offering a numerical assessment of the overall efficacy of ablation.

This pioneering system compensates for complex lung deformations and integrates pre- and intra-operative imaging data, enhancing quality control in cancer ablation treatments. A follow-up study involving 35 clinical cases demonstrated that our system significantly outperforms traditional subjective assessments in identifying under-ablation cases during or immediately after treatment, highlighting its potential to improve clinical decision-making and patient outcomes.
\end{abstract}

\section{Introduction}

%Lung cancer remains one of the most significant global health challenges, consistently ranking as the leading cause of cancer-related mortality worldwide. According to the World Health Organization (WHO), an estimated 2.2 million new cases and 1.8 million deaths were reported in 2020 alone \cite{sung2021global}. The disease is characterized by malignant tumors originating from various cell types within the pulmonary structure \cite{chi2021technical}. Surgical resection is the primary treatment modality recommended by the National Comprehensive Cancer Network (NCCN), involving the excision of both tumor and adjacent lung tissue \cite{cazzato2015cone}. However, this approach often requires the removal of substantial amounts of healthy lung tissue, which is unnecessary for the treatment of small, early-stage tumors \cite{chen2017assessments,nance2021computerized}. Factors such as patient age, comorbidities, or reluctance to undergo invasive procedures often preclude surgical intervention \cite{cykert2010factors}.

In recent years, image-guided thermal ablation techniques, including RadioFrequency Ablation (RFA) and MicroWave Ablation (MWA), have emerged as minimally invasive alternatives. These methods offer advantages such as reduced recovery time and lower procedural risks compared to traditional surgery \cite{lin2020image,prud2019image,ni2020image}. Among these, MWA is particularly effective in the pulmonary environment as a result of its superior control over the ablation zone, enabled by the low electrical conductivity of the lungs. Extensive research supports the efficacy and safety of MWA for the treatment of lung tumors \cite{yang2014percutaneous,yang2018microwave,zhong2017clinical,yuan2019meta}. The procedure involves the percutaneous insertion of a CT-guided ablation probe into the tumor, where thermal energy is used to destroy cancer cells.

Despite MWA's precision in targeting tumor tissue while minimizing damage to healthy lung tissue, residual tumor risks remain a significant concern. This challenge arises from the difficulty of accurately assessing whether ablation fully encompasses the tumor area. Post-ablation evaluation typically involves comparing preoperative and postoperative CT scans, but respiratory-induced lung deformation and ablation-related imaging changes complicate accurate alignment. Consequently, physicians often rely on subjective judgment to determine the completeness of the ablation, which increases the risk of overestimating treatment success.

To address these challenges, we propose \textit{Respiratory Differencing}, a novel intraoperative CT image assistance system designed to enhance the evaluation of pulmonary thermal ablation. This system addresses two critical needs for physicians during or immediately after the procedure. First, it allows direct visual comparison between the tumor and treatment areas, despite the complexity of lung deformation during CT/CBCT imaging at different time points, as well as the challenge that the tumor in the preoperative image may have been destroyed or significantly altered in the intraoperative/postoperative images. Second, the system introduces a quantitative evaluation metric, the \textbf{A}blation \textbf{E}ffectiveness \textbf{S}cale (AES), which measures discrepancies between the intended tumor areas and the treatment zones. This metric provides an objective assessment of ablation efficacy, aiding in systematically evaluating and optimizing treatment protocols.

The system utilizes multi-stage deformable lung registration techniques to align preoperative and intra/postoperative CT images, compensating for respiratory-induced deformations and treatment-related changes. The process begins with tumor segmentation on preoperative images and identification of the treatment zone on postoperative images. These segmented images are then aligned using a multi-stage registration approach, which involves coarse rigid registration followed by fine non-rigid registration. This approach accounts for variations in respiratory phases, ensuring accurate alignment of tumor areas and treatment zones.

Once the images are aligned, the system generates differential images by subtracting the registered preoperative images from the postoperative ones. These \textit{Respiratory Differencing} images provide clear visualizations of the changes, enabling a precise and accurate assessment of ablation completeness. Furthermore, the AES metric quantifies the discrepancies between the intended tumor areas and the treatment zones, offering an objective evaluation of the ablation process and supporting the optimization of treatment protocols. Physicians can rely on the visual output of \textit{Respiratory Differencing} or the evaluation score to quality control the effect of the treatment and identify incomplete treatment cases.

%To address these limitations, we propose \textbf{Respiratory Differencing}, a novel intraoperative CT image assistance system for enhanced evaluation of pulmonary thermal ablation. This system leverages deformable lung registration techniques to align preoperative and intra/postoperative CT images, compensating for respiratory-induced deformation and ablation-related imaging changes. The process begins with tumor segmentation in preoperative images and treatment zone identification in postoperative images. These segmented images are aligned using multi-stage registration, including coarse rigid and fine non-rigid registration, to unify respiratory phases and ensure accurate alignment of tumor areas and treatment plans.

%Once aligned, the system generates differential images by subtracting registered preoperative images from postoperative ones. These "respiratory differencing" images allow for direct visualization of changes, enabling a more objective assessment of ablation completeness. Additionally, we introduce a quantitative evaluation metric, the Ablation Effectiveness Scale (AES), to measure discrepancies between intended tumor areas and treatment zones. This metric provides an objective assessment of ablation efficacy, supporting systematic evaluation and optimization of treatment protocols.

To validate the efficacy of the proposed system, we conducted a comparative analysis of system-generated AES scores and subjective assessments by physicians. A follow-up study involving 35 clinical cases demonstrated that our system outperformed traditional methods in identifying instances of under-ablation. By improving the precision of the assessment, our system has the potential to improve clinical decision making, reduce incomplete tumor treatment, and improve patient outcomes.

In summary, this paper:
\begin{enumerate}
    \item[(a)] Identifies that more than 43\% of the cases initially assessed as over-ablation during or immediately after treatment were later found to be under-ablation or normal ablation upon follow-up. This finding aligns with previous studies \cite{granata2021diagnostic} and underscores the limitations of current empirical assessment methods.
    \item[(b)] Develops a CBCT/CT-CT image registration system that provides fused preoperative and intraoperative images, allowing direct spatial comparisons of tumor and treatment zones on a unified graphical interface.
    \item[(c)] Introduces the Ablation Effectiveness Scale (AES), a quantitative metric for evaluating ablation efficacy based on image registration and automatic segmentation.
\end{enumerate}

Through rigorous clinical data testing, we demonstrate the effectiveness of our image registration system and the AES index. In particular, adopting our system increases Spearman's correlation coefficient between immediate postoperative imaging and subsequent medical records from 0.443 to 0.809, surpassing traditional physician assessments based on empirical observation.

%This finding echoes other related studies and highlights the necessity of graphical tools to help physicians compare the relative spatial relationship between cancer and treatment area 
%This finding echoes other related studies and highlights the necessity of graphical tools to help physicians compare the relative spatial relationship between cancer and treatment area 
%This finding echoes other related studies and highlights the necessity of graphical tools to help physicians compare the relative spatial relationship between cancer and treatment area and ultimately provide quantitative evaluation of 
%that more than 85\% cases that are considered to be over-ablation during or immediately after treatment is actually average under a 
%In summary, Respiratory Differencing represents a significant advancement in the evaluation of pulmonary thermal ablation by providing a reliable, quantitative framework for treatment assessment. By overcoming the challenges associated with lung deformation and integrating preoperative and postoperative imaging information, our system offers a robust tool for enhancing quality control in cancer ablation treatments, paving the way for more effective and controlled therapeutic interventions.

\section{{Treatment Evaluation}}

To examine the precision of the current clinical protocol in evaluating the completeness of lung cancer ablation treatment, two highly skilled physicians specializing in lung ablation surgery were invited to independently rate subjective scores for CT images from all patients at different time points: immediately postoperative, 1 month postoperative, 3 months postoperative, and 6 months postoperative. The ratings of CT images taken immediately after surgery and during follow-up (paired with all medical records) were used to compare the correlation with our AES index metric. Additional postoperative follow-up medical records, such as images showing proliferating tumors when initial ablation surgery failed, were also considered. The physicians assigned subjective scores classified into three levels corresponding to common ablation surgery outcomes: 1 for overablation, 0 for average ablation, and -1 for underablation. The scores collected showed a high Kappa value of 0.645 (p\textless 0.001), indicating a high level of consistency in the physician evaluations.

% \begin{table}[]
%     \centering
%     \caption{Physicians' assessments over the same patient cohort based on intraoperative CT images and 6 months postoperative images}
%     \begin{tabular}{lccc}
%         \hline
%         \textbf{Date\textbackslash Assessment} & \textbf{Under} & \textbf{Average} & \textbf{Over} \\
%         \hline
%         \textbf{Intraoperative CT images} & 5  & NaN   & 30\\
%         \textbf{Long-term postoperative CT images} & 3 & 14  &  18 \\
%         \hline
%     \end{tabular}
%     \label{intra_post_compare}
% \end{table}

\begin{table}[]
    \centering
    \caption{Confusion matrix of physicians' assessments  between immediately and 6 months postoperative}
    \begin{tabular}{lccc}
        \hline
        \textbf{Immediately\textbackslash 6 Months}& \textbf{Under} & \textbf{Average} & \textbf{Over} \\
        \hline
        \textbf{Under ablation} & 2 & 2 & 1 \\
        \textbf{Average ablation} & 0 & 0 & 0 \\
        \textbf{Over ablation} & 1 & 12 & 17 \\
        \hline
    \end{tabular}
    \label{confusion_doc}
\end{table}

% Table \ref{intra_post_compare} illustrates the class distribution between physicians' assessments. 
Longer-term postoperative ratings paired with medical records can be considered as the ground truth assessment since it was based on all available information and the subsequent results of the therapy. The overproportion of the over-ablation labels in the intraoperative part of the table shows that, on intraoperative/immediate-postoperative images, doctors tend to overestimate ablation ratings and many cases that should be labeled as under-ablation or average-ablation were miss-labeled as over-ablation. The possible reason is that by simply looking at the immediate treatment image, doctors may not be able to remember the exact location of the preoperative cancer area due to lung deformations. Therefore, they may tend to assess the treatment primarily on the basis of the absolute ablation area. 

Table \ref{confusion_doc} shows the confusion matrices between the physicians' subjective scores immediately after surgery and 6 months after. In Table \ref{confusion_doc}, we can find that more than 43\% of the cases judged as over-ablation by doctors immediately after surgery were later rated as under-ablation or average-ablation. This proves that doctors tend to overestimate the ablation ratings when merely judging from the images immediately after treatment. This phenomenon suggests that knowing the exact location of the original cancer in postoperative images is vital to prevent misjudgments. Our method provides both image and score to assist the doctors so that they can better understand the relative position between the postoperative ablation area and the preoperative cancer area immediately after surgery. These cases also demonstrate the importance of our registration system, which can provide the exact location of cancer even after it disappears after ablation. Our \textit{Respiratory Differencing} image could give doctors a solid reference point when assessing whether ablation treatment is sufficient and, therefore, prevents misjudgment.

%\begin{figure}
%\centering
%\includegraphics[width=3.5in]{class-distribution-lung-updated1.png}
%\caption{The distribution of doctors' subjective rating on images \textcolor{blue}{after surgery and at 6 months postoperative.}}
%\label{data_distribution}
%\end{figure}

%As shown in Table \ref{spearman}, the Spearman correlation analysis results indicate a robust positive correlation between the AES index and the doctors' ratings (6 months after surgery), which suggests that the AES index can provide valuable support in assessing clinical ablation procedures. Compared with the Spearman correlation coefficient value of 0.443 from the immediate postoperative result, the value of 0.788 indicates that our method shows a stronger correlation with the accurate long-term judgment results. Meanwhile, Table \ref{confusion} shows a confusion matrix between the doctors' subjective scores 6 months after and our AES index (thresholded by $\lambda$). From table \ref{confusion} we can see that our method can predict the under-ablation cases with full precision, with all the under-ablation cases in the experiment dataset accurately found and rated. As mentioned, under-ablation scenarios can potentially cause dangerous mistreatment or readmission to excessive therapies afterwards. This result shows that our method could preemptively prevent such possibilities caused by doctors' overlook of under-ablation scenarios and protect patients from missing the best treatment opportunity or being subjected to unnecessary second ablation therapy afterwards.

\section{Method}

\begin{figure}[t]
\centering
\includegraphics[width=4.6in]{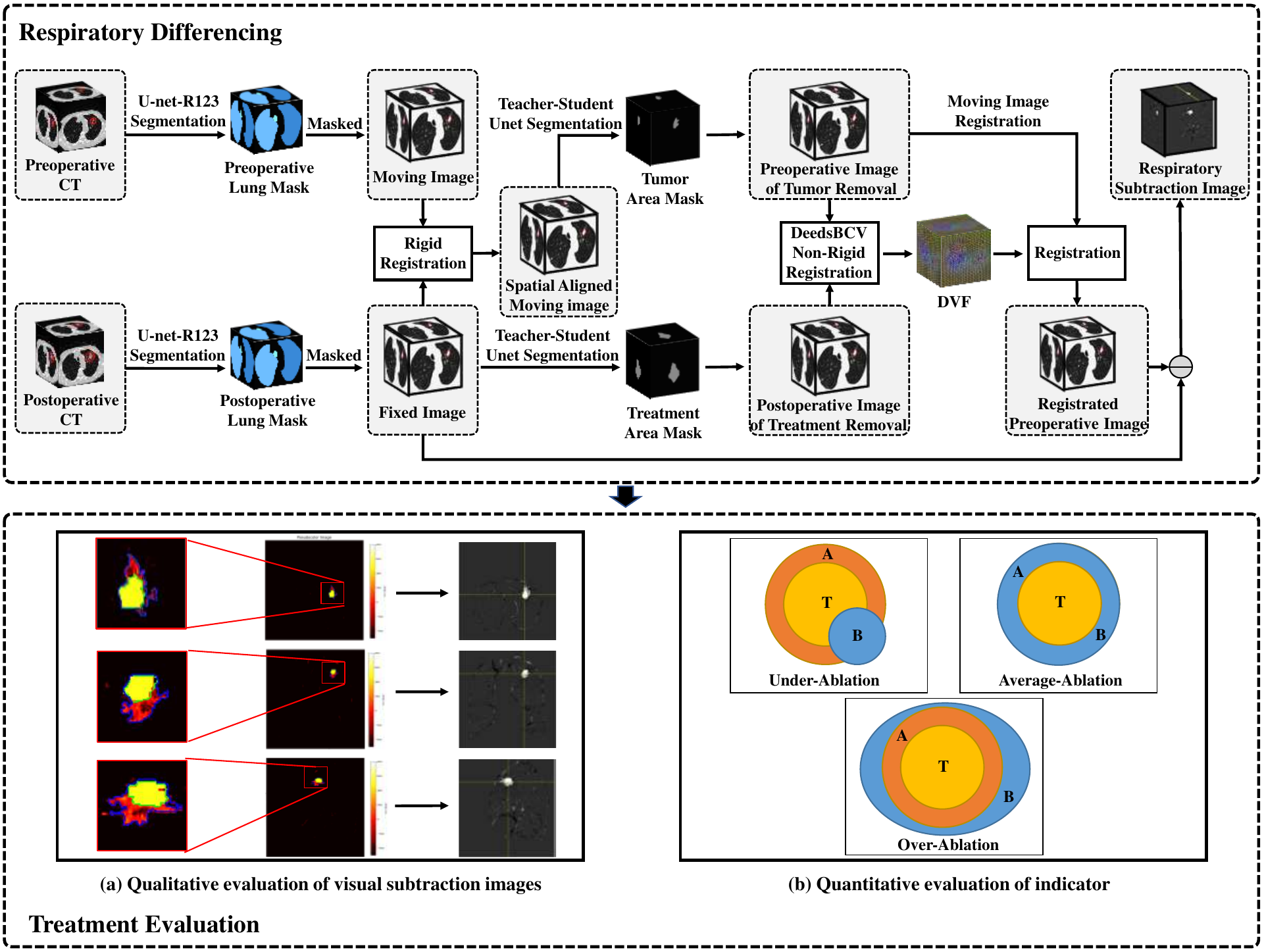}
\caption{The overall framework for \textit{Respiratory Differencing} and AES quantitative assessment.}
\label{fig:framework}
\end{figure}

The general framework of our treatment evaluation system is shown in Figure \ref{fig:framework}. Lung parenchymas are segmented from preoperative and intraoperative/postoperative images. The preoperative images are then registered to the intraoperative/postoperative images. The \textit{Respiratory Differencing} image is acquired by subtracting the preoperative and intraoperative/postoperative images. The \textit{Respiratory Differencing} images are subsequently used in subjective evaluations of surgical effectiveness and in the computation of the AES score.

\subsection{Lung Parenchyma Segmentation}

For processing pulmonary MWA data, registration to reduce respiratory motion is accomplished by extracting the lung parenchyma separately. This approach is adopted because the clinical focus is solely on the lung parenchyma, and it improves the segmentation performance of tumors and treatment areas, leading to more accurate assessments. Specifically, Unet-R231\cite{Hofmanninger2020AutomaticLS}, an advanced Unet-based structure known for its superior lung segmentation capabilities, is the first step of our method. This model generates lung parenchyma masks for preoperative and postoperative images. Training in a routine clinical dataset (n = 231) containing multiple diseases, the model provides masked images $P_{pre}$ and $P_{post}$. These masked images serve as the moving and fixed images, respectively, for subsequent rigid registration of the lung parenchyma.

\subsection{Lung Parenchyma Rigid Registration}

To achieve an initial alignment between preoperative and postoperative lung images, we perform rigid registration on the extracted lung parenchyma images \(P_{\text{pre}}\) and \(P_{\text{post}}\). This process applies rotations and translations to \(P_{\text{pre}}\) to minimize differences, quantified using Normalized Cross-Correlation (NCC) as a similarity measure:

% The initial coarse registration of the lung parenchyma is then performed to ensure the general placement and alignment of the preoperative and postoperative images. The rigid three-dimensional registration is carried out on the extracted preoperative and postoperative lung parenchyma images $P_{pre}$ and $P_{post}$ applying rotations and translations to the moving image $P_{pre}$. In order to minimize the differences between $P_{pre}$ and $P_{post}$, the alignment process is quantified using a loss function Normalized Cross-Correlation (NCC). NCC measures the similarity between the two images and is defined as:

%\begin{equation}
%\text{NCC}(P_{\text{pre}}, P_{\text{post}}) =
%\frac{\displaystyle\sum_{i,j,k} \left( P_{\text{pre}}(i,j,k) - \bar{P}_{\text{pre}} \right) \left( P_{\text{post}}(i,j,k) - \bar{P}_{\text{post}} \right)}
%{\displaystyle\sqrt{\sum_{i,j,k} \left( P_{\text{pre}}(i,j,k) - \bar{P}_{\text{pre}} \right)^2 \sum_{i,j,k} \left( P_{\text{post}}(i,j,k) - \bar{P}_{\text{post}} \right)^2}}
%\end{equation}

%where \( \bar{P}_{\text{pre}} \) and \( \bar{P}_{\text{post}} \) are the mean intensities of the preoperative and postoperative images, respectively. The rigid registration aims to find the transformation \( T_1 \) that maximizes the NCC value, indicating the best alignment of the images.

The optimal rigid transformation \(T_1\), consisting of a rotation matrix \(R\) and a translation vector \(\mathbf{t}\), is iteratively optimized to maximize NCC. We employ SimpleITK-based algorithms for this optimization, ensuring precise alignment before proceeding to non-rigid registration.
The resulting rigid transformation $T_1$ can be mathematically described as:
\begin{equation}
T_1(\mathbf{x}) = R \mathbf{x} + \mathbf{t}
\end{equation}
% where \( R \) is a \(3 \times 3\) rotation matrix, \( \mathbf{t} \) is a translation vector, and \( \mathbf{x} \) represents the coordinates of the points in the image. The transformation \( T_1 \) is iteratively optimized to align \( P_{\text{pre}} \) with \( P_{\text{post}} \). This optimization is performed using algorithms available in the SimpleITK library, which adjust the parameters of \( R \) and \( \mathbf{t} \) to improve the similarity measure given by the NCC. After this step, the lung parenchyma from the preoperative images is adjusted to the position of the lung parenchyma from the postoperative images, preparing for the non-rigid registration of lung tissues.

\subsection{Tumor Segmentation and Treatment Area Segmentation}

Before non-rigid registration of lung tissues, it is necessary for preprocessing to segment and remove tumors from the preoperative images and treatment areas from the postoperative images. Retaining these regions could cause them to be interpreted as prominent features, adversely affecting the non-rigid registration process. To minimize such interference, tumor region masks $M_{pre}$ are extracted from preoperative images $P_{pre}$, and treatment area masks $M_{post}$ are extracted from the postoperative images $P_{post}$.
These masks are automatically generated using a Unet-based architecture trained by the teacher-student approach\cite{Fredriksen2022} and then the corresponding part is excluded from the images.

\subsection{DeedsBCV Non-Rigid Registration}

%To account for lung tissue deformation between preoperative and postoperative stages, we employ the non-rigid registration method deedsBCV\cite{6471238}. This approach uses a hierarchical, multi-resolution strategy to compute dense displacement fields, refining both large- and small-scale deformations. The optimization is formulated as an energy minimization problem, balancing data fidelity and smoothness constraints. Unlike traditional gradient-based methods, deedsBCV operates in a discrete label space, reducing the risk of local optima and improving registration accuracy. 

 Taking into account the characteristics of lung tissues and the possible significant deformations in the lung areas between the preoperative and postoperative stages, a non-rigid registration method, deedsBCV\cite{6471238}, is introduced to further fine-tune the registration results. Deeds in deedsBCV stands for DEnsE displacement sampling. It applies a hierarchical, multi-resolution approach to calculate dense displacement fields, gradually optimizing at different resolution levels to handle both large-scale and small-scale deformations in the images. This method constructs a graph model using control points in a uniform B-spline grid as nodes, with an optimized energy function composed of a data term and regularization term:

 \begin{equation}
 E(f) = \underbrace{\sum_{p \in P} D(f_p)}_{\text{data term}} + 
 \alpha \underbrace{\sum_{(p,q) \in N} R(f_p, f_q)}_{\text{regularization term}}.
 \end{equation}

 where $p$ denotes the control points corresponding to the spatial location $x_p$, and for each node, $f_p$ is the discrete displacement. The node $q$ denotes the direct neighbor node connected to $p$. $\alpha$ is a hyperparameter that controls the balance between the data term and the regularization term. The data term describes the similarity between the control point in one image and the corresponding control point in another image with discrete displacement $f_p$. In our implementation, the self-similarity descriptor is adopted to calculate the similarity of the corresponding nodes. The pairwise regularization term ensures global smooth transformation by penalizing the squared displacement differences of adjacent control points:
 \begin{equation}
 R(f_p, f_q) = \sum_{(p, q) \in {N}} \frac{\left\| \mathbf{u}_p - \mathbf{u}_q \right\|^2}{\left\| \mathbf{x}_p - \mathbf{x}_q \right\|}
 \end{equation}

 A significant advantage of deedsBCV is its discrete optimization framework. Unlike traditional gradient-based optimization methods, deedsBCV operates within a discrete label space. Each voxel in the image is assigned a discrete displacement label, representing potential deformations at that point. This discrete approach helps to avoid common local optima found in continuous optimization, thereby achieving more accurate registration results.

After the non-rigid registration, we get the non-rigid deformation field $T_2$. The final registered preoperative image $P_{registered}$ can be obtained using the equation
\begin{equation}
P_{registered}=P_{pre} \circ T_1 \circ T_2
\end{equation}
where $\circ$ represents the registration operation.

\subsection{Differencing}

After aligning the preoperative and postoperative images, the \textit{Respiratory Differencing} images $P_{sub}$ can be easily obtained by subtracting the registered preoperative images from the postoperative images:
\begin{equation}
P_{sub}=P_{post} - P_{registered} = P_{post} - P_{pre} \circ T_1 \circ T_2
\end{equation}

The resulting differencing image $P_{sub}$ offers enhanced visual confirmation of lung MWA treatment performance compared to traditional observation methods, and visual evaluation can be performed by comparing the tumor area and treatment limit in $P_{sub}$. Additionally, quantitative evaluation can be achieved and will be introduced in our next subsection.

\subsection{AES index}

To further assist doctors in evaluating surgery, in addition to the visualization aid provided by the differencing image $P_{sub}$, we propose a new quantitative metric based on $P_{sub}$ to evaluate the effectiveness of lung MWA treatment, named the Ablation Effectiveness Scale (AES) index.

An effective treatment should include the tumor area within the treatment area to ensure complete tumor ablation, but it should not extend excessively beyond, which can cause damage to surrounding healthy tissues. The treatment boundary should aim to minimize the treatment area while still achieving complete tumor ablation. Clinically, it is generally recommended that the ablation range extends 5-10mm beyond the tumor border to ensure the edges of the tumor are treated and to reduce the risk of residual lesions.

\begin{figure}[]
\centering
\includegraphics[scale=0.55]{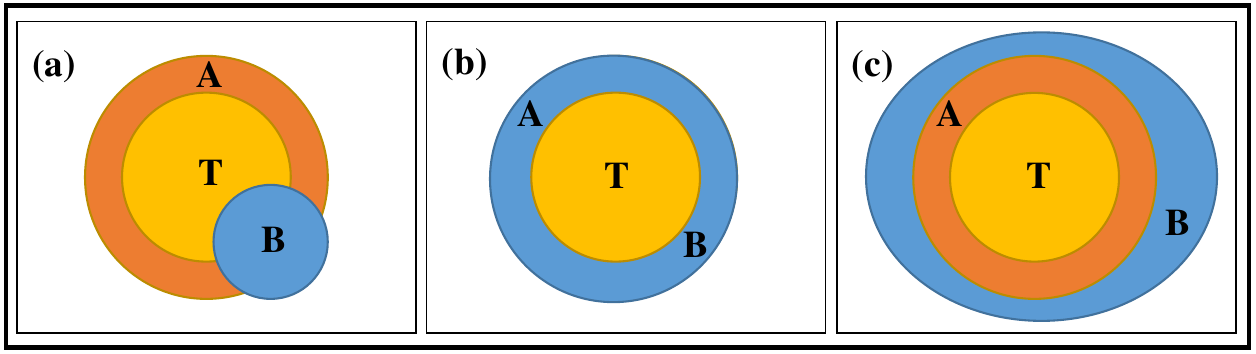}
% \captionsetup{justification=raggedright, singlelinecheck=false}
\caption{Respiratory Differencing ablation therapy evaluation
    (a) Under-ablation; (b) Average-ablation; (c) Over-ablation; 
    $T$: Tumor; $B$: Actual ablation; $A$: Planned ablation.}
\label{figure8}
\end{figure}

In clinical scenarios, the outcomes of ablation treatment can be broadly categorized into three types: under-ablation, average-ablation (or ideal ablation), and over-ablation (or excessive ablation), as shown in Figure \ref{figure8}. Under-ablation refers to the situation where the tumor area is not completely covered by the ablation area, indicating surgical failure and potentially necessitating an additional ablation surgery for the patient. Average-ablation indicates that the tumor area is completely covered by the ablation area and that the ablation area is within the optimal range for the procedure. Over-ablation means that while the tumor area is completely covered, the ablation area exceeds the clinically acceptable range, potentially causing unnecessary damage to normal tissues.

To better reflect the therapeutic outcomes of ablation surgery, our metric, the AES index, is designed to assess potential scenarios in lung ablation surgery from three aspects: actual treatment tumor coverage, planned ablation area coverage, and the degree of over-ablation. To facilitate understanding, the symbols and their meanings in the subsequent formulas are defined as follows. The tumor volume and the ablation volume are denoted as $T$ and $B$, respectively. The planned ablation volume of the tumor is denoted as $A$, which is obtained by expanding the boundary of volume T by 5mm outward. The symbols $\cap$ and $\setminus$ represent the intersection and difference of two sets, respectively. Based on these definitions, the AES index can be calculated using Eq \ref{eq8} to evaluate the effectiveness of lung MWA surgery:

\small
\begin{equation}
CR_1=\frac{T\cap B}{T},\ \  
CR_2=\frac{B\cap A}{A},\ \ 
ER=\frac{B\setminus A}{B}\label{eq7}
\end{equation}
% \small
% \begin{eqnarray}
% CR_1=\frac{T\cap B}{T}\label{eq5}
% \end{eqnarray}
% \small
% \begin{eqnarray}
% CR_2=\frac{B\cap A}{A}\label{eq6}
% \end{eqnarray}
% \small
% \begin{eqnarray}
% ER=\frac{B\setminus A}{B}\label{eq7}
% \end{eqnarray}
\small
\begin{equation}
AES = \left\{
\begin{array}{ll}
-1+ e^{- \alpha \cdot (1-CR_1) - \beta \cdot ER} & \text{if } CR_1 < \lambda, \\
1 - e^{- \gamma \cdot (1-CR_2) - \theta \cdot ER} & \text{if } CR_1 \geq \lambda
\end{array}
\right. \label{eq8}
\end{equation}

where $CR_1$ denotes the current tumor coverage ratio of the ablation area, $CR_2$ is the ratio of the current valid ablation area to the planned ablation area and $ER$ is used to represent the degree of over-ablation. The AES index is specifically defined in Eq \ref{eq8}. In the case of low tumor coverage ($CR_1 < \lambda$), the primary focus of the AES index lies in ensuring complete tumor coverage and evaluating the proportion of the invalid ablation area. Empirically, the hyperparameters $\alpha$ and $\beta$ are set at 1.0 and 2.0, respectively, to balance tumor coverage and the rate of over-ablation in subsequent experiments. As tumor coverage increases in this scenario, the AES index value also increases gradually toward zero. Although at a similar level of tumor coverage, a higher degree of over-ablation results in a lower AES index value. When tumor coverage exceeds the threshold $\lambda$, the AES index primarily emphasizes ablation area coverage and the proportion of over-ablation. Similarly, the hyperparameters $\gamma$ and $\theta$ are empirically set to 0.5 and 2.0, respectively, to strike a balance between these factors. In the second scenario, as the ablation area coverage increases, there is a decrease in the corresponding AES index value. While at an equivalent level of ablation area coverage, less over-ablation leads to smaller values approaching zero. In our experiment, the parameter $\lambda$ is empirically set to 0.75.

\section{Experiment and Results}

\subsection{Patient Cohort}

A monocentric retrospective study was conducted between April 2022 and November 2023 in Beijing Chaoyang Hospital, Capital Medical University (Chinese Clinical Trial Registry, ID: ChiCTR2300076517). Written informed consent was obtained from all participating patients before they underwent CT-guided percutaneous MWA or bronchoscopic MWA of their lung tumors. Patients were recruited based on the following eligibility criteria: (1) age \(\geq\)18 years old; (2) malignant lung tumors confirmed by histological or cytological diagnosis; (3) patients who cannot tolerate surgery and stereotactic body radiation therapy (SBRT) due to older age, poor cardiopulmonary function, or refusal of surgery and SBRT; (4) the diameter of the target lesion was \(\leq\)3 cm, with lesion size defined as the longest diameter on the largest cross-section of the CT lung window; (5) re-examination by chest CT at 1 day, 1 month, 3 months, and 6 months after the procedure; (6) exclusion of pregnancy, coagulopathy, severe mental illness, severe hypoxemia, hemodynamic instability, or pacemaker installation. A cohort of 35 patients who had received MWA for a lung tumor were identified for study inclusion.

The ages of the patients range from 41 to 87 years, with 18 males and 17 females represented. The lesion areas range from 11.2 mm² to 570 mm², with an average lesion area of approximately 176.6 mm². Before the intervention, all patients underwent an enhanced plain chest CT examination in 4 weeks. The surgical approach of MWA was determined based on the imaging characteristics of the lung lesions. Navigation bronchoscopy was used for ablation when the pulmonary lesion has a positive bronchial sign or is adjacent to the bronchial, and otherwise, CT-guided percutaneous was used. The ablation power ranges from 10W to 60W, with durations of 2 to 15 minutes.

\begin{figure}
\centering
\includegraphics[width=3.5in]{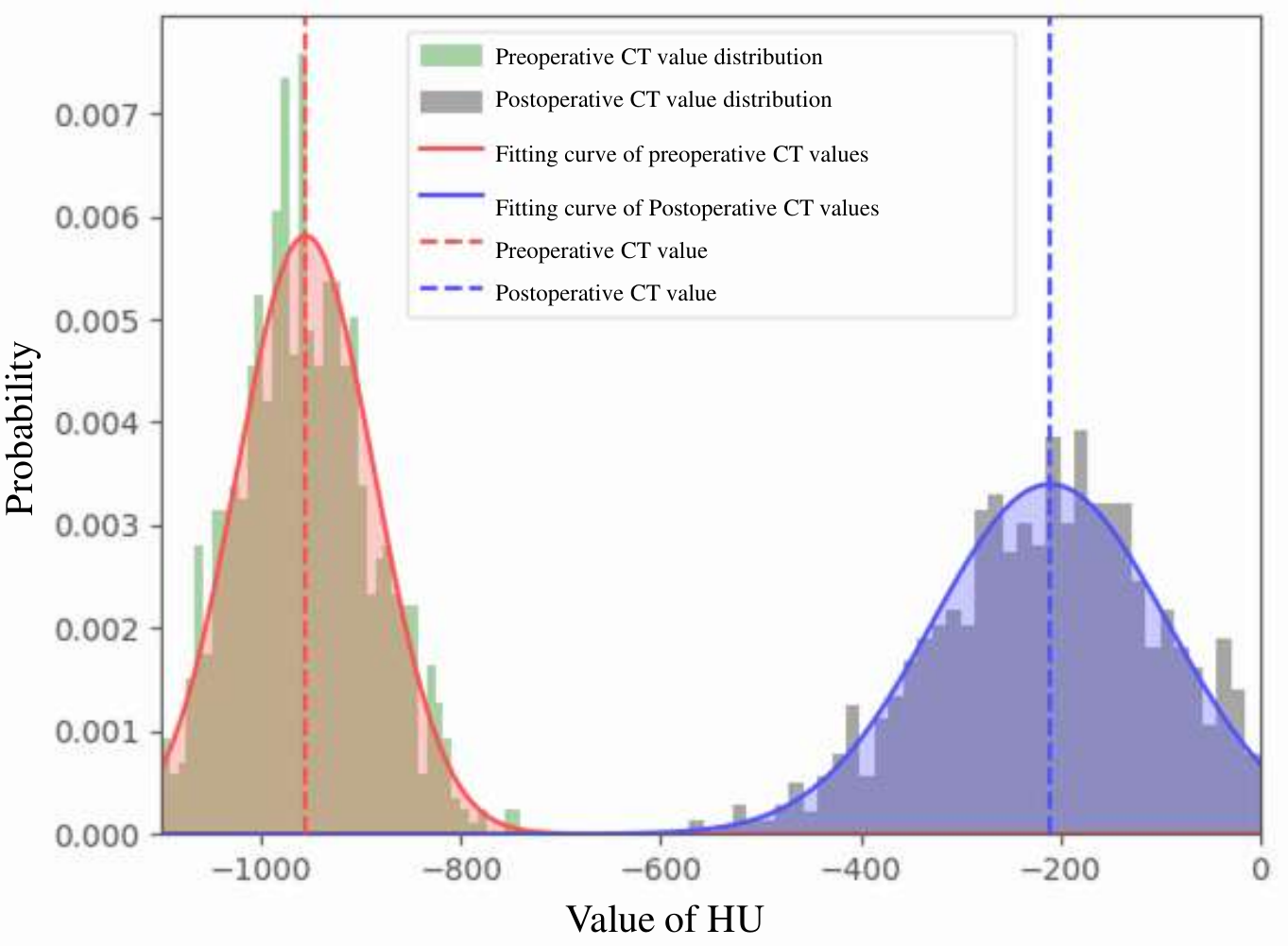}
\caption{HU distribution comparison between the preoperative tumor area and postoperative ablation treatment area.}
\label{fig_5}
\end{figure}

\subsection{Dataset}
% In this study, clinical preoperative and postoperative CT images were collected at the time of immediate postoperative, 1 month post-surgery, 3 months post-surgery, and 6 months post-surgery from 35 patients at Beijing Chao Yang Hospital, and written informed consent was provided by all patients. 
The majority of the CT images used in this study are taken by GE MEDICAL SYSTEMS's Revolution CT (General Electric Company, USA) with a peak kilo voltage of 120kV. In our experiment, all the involved ground truth masks were manually annotated by doctors. All the CT data underwent preprocessing, which included cropping to the dimension of 256*256*256 and configuring the spacing to 1.25mm*1.25mm*1.25mm. Subsequently, normalization was performed to scale the pixel values of the data within the range between zero and one. The proposed \textit{Respiratory Differencing} method and the aforementioned evaluation metrics were utilized in the conducted experiments. For all statistical analyses, SPSS software was used to perform Kappa analysis and Spearman correlation analysis.

\begin{figure}[t]
\centering
\includegraphics[width=5in]{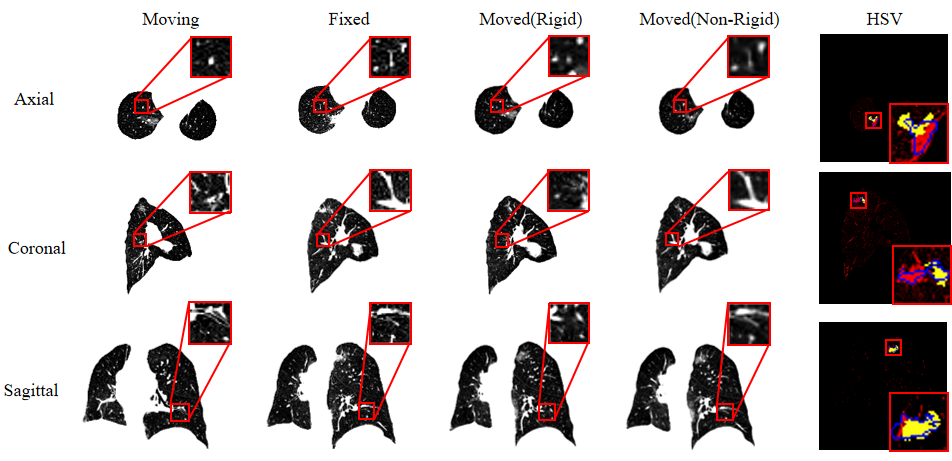} 
\caption{Registration results after tumor segmentation and corresponding HSV pseudo-color images on the real clinical data. The HSV pseudo-color images represent the difference between the registered moved images and the original postoperative images. Preoperative tumor masks after registration are remapped onto the HSV images and displayed in yellow, while the postoperative treatment area boundaries are depicted in blue. Red regions in these HSV images represent significant changes in HU values within the ablation area before and after the procedure. The image depicts a \textit{Respiratory Differencing} example of an unsuccessful ablation procedure of a patient, which has been confirmed by medical records that this patient underwent surgical intervention for the tumor again after 3 months. It can be evidently observed that the ablation area (blue) did not completely encompass the tumor (yellow).}
\label{fig_4}
\end{figure}

\subsection{Statistical analysis of pre- and post-operative HU around tumor region}

The Hounsfield Unit (HU) values around the tumor region were statistically analyzed both preoperative and postoperative, as depicted in Figure \ref{fig_5}. The analyzed region specifically refers to the area outside the tumor area but within the treatment area. The statistical analysis reveals a significant difference in HU values in the region before and after surgery, with the two distributions distinguishable using a threshold of around -600 HU. This statistical finding has significant clinical implications for the evaluation of lung ablation surgeries. It allows for determining the range of ablation by assessing the average HU value at the ablation site. The finding aligns with the clinical follow-up outcomes obtained from 35 patients.

\subsection{Registration results for Respiratory Differencing}

To demonstrate the efficacy of \textit{Respiratory Differencing}, we conducted experiments on the collected dataset and visualized the differencing results as HSV pseudo-color images. as depicted in Figure \ref{fig_4}. The figure illustrates that rigid registration can roughly align the images, while non-rigid registration precisely aligns the lung tissues in detail. By comparing the images in the "Moved (Non-Rigid)" column with those in the "Fixed" column, it is evident that the detailed lung tissue structures in the moved (preoperative) images are well-aligned with those in the fixed (postoperative) images. The findings indicate that the proposed \textit{Respiratory Differencing} method achieves highly accurate registration results in critical soft tissues of the lungs, demonstrating precise mapping of fundamental lung features to their corresponding positions.

\begin{table}[t]
\caption{Quantitative comparison of clinical data registration results of different registration methods}
\centering

\begin{tabular}{c|cccc}
        \hline
            & \textbf{NCC $\uparrow$}  & \textbf{SSIM  $\uparrow$} & \textbf{RMSE $\downarrow$}  & \textbf{Dice $\uparrow$}  \\
        \hline
        \textbf{No Registration}  & 0.6301 & 0.8040 & 230.76 & 0.7015 \\
        \textbf{MI Rigid Registration}     & 0.8805 & 0.8567 & 131.04 & 0.9002 \\
        \textbf{FSFDRF} & 0.9398 & 0.8985 & 89.09  & 0.9478\\
        \textbf{Our method} & \textbf{0.9733} & \textbf{0.9231} & \textbf{59.34}  & \textbf{0.9791}\\
        \hline
    \end{tabular}
    \label{tab1}
\end{table}

The precision of the \textit{Respiratory Differencing} registration process was further quantitatively analyzed using Normalized Cross-Correlation (NCC), Structural Similarity (SSIM), Root Mean Square Error (RMSE) and Dice coefficients for lung parenchyma masks derived from postoperative and preoperative images registered.   Due to the narrow field of view in intraoperative imaging and interference from surgical instruments and pipelines,  deep learning algorithms trained on public datasets showed inferior performance compared to traditional algorithms in our study. Therefore, we choose to compare our registration algorithm with rigid registration based on Mutual Information (MI) and the Fast-Symmetric-Forces-Demons-Registration Filter (FSFDRF)\cite{ITK}. The rigid registration algorithm based on mutual information is remarkably effective in aligning multi-modal medical images, showcasing exceptional robustness and precision in clinical scenarios. In parallel, the FSFDRF algorithm excels in handling complex deformations and anatomical variances, making it suitable for scenarios such as lung ablation surgery. In comparison, our algorithm, based on deedsBCV, is less prone to local optima than FSFDRF and is better suited for potential multimodal scenarios (such as CT/CBCT) before and after surgery, effectively mitigating the interference of the surgical instruments and narrow field of view when registering intra-operative images.

Table \ref{tab1} presents the results of these registration metrics on the experiment's clinical data. The results demonstrate that our method achieved a significant improvement in Dice coefficients compared with both MI registration and FSFDRF registration, all with values exceeding 0.90. Among these methods, our algorithm achieved the highest Dice coefficient of 0.98, indicating superior registration performance. Furthermore, there was a notable decrease in RMSE, which also indicated that our registration process effectively aligns overall lung regions. In addition, notable improvements were observed in NCC and SSIM, suggesting that our registration process effectively aligns local lung details to meet clinical application requirements and provides solid performance for clinical decision-making and evaluation in pulmonary MWA surgery.

\begin{table}[t]
    \centering
    \caption{{Correlation between AES index (immediately postoperative) and physicians' assessments  (immediately and 6 months postoperative)}}
    \begin{tabular}{lcc}
\hline
\textbf{Metric}               & \textbf{Immediately}  & \textbf{6 Months After}\\
\hline
\textbf{Spearman correlation}&     0.443       & 0.809\\
\textbf{P-value}              &     7.761E-3& 3.996E-9\\
\hline
\end{tabular}
\label{spearman}
\end{table}

\begin{table}[t]
    \centering
    \caption{{Confusion matrix between 6 months postoperative physicians' assessments  and AES index}}
    
\begin{tabular}{lccc}
\hline
\textbf{6 Months\textbackslash AES}& \textbf{\textless $-\lambda$} &  \textbf{$\sim$0} & \textbf{\textgreater $\lambda$} \\
\hline
\textbf{Under ablation} & 5.71\%& 2.86\%& 0.00\%\\
\textbf{Average ablation}  & 0.00\%& 37.14\%& 2.86\%\\
\textbf{Over ablation}  & 0.00\% & 5.71\%& 45.71\%\\
\hline
\end{tabular}
\label{confusion}
\end{table}

\subsection{Validity analysis of AES index}

To validate the effectiveness of the AES index in evaluating ablation surgery efficacy, two highly skilled physicians specializing in lung ablation surgery were invited to independently rate subjective scores for all patients' CT images at different time points: immediately postoperative, 1 month postoperative, 3 months postoperative and 6 months postoperative. The ratings of CT images taken immediately after surgery and during follow-up (paired with all medical records) were used to compare correlation with our AES index metric. Additional postoperative follow-up medical records, such as images showing proliferating tumors when initial ablation surgery failed, were also considered. Physicians assigned subjective scores classified into three levels corresponding to common ablation surgery outcomes: 1 for over-ablation, 0 for average ablation, and -1 for under-ablation. The collected scores demonstrated a high Kappa value of 0.645 (p\textless 0.001), indicating a high level of consistency in the doctors' assessments.

\begin{figure}[t]
\centering
\includegraphics[width=3.5in]{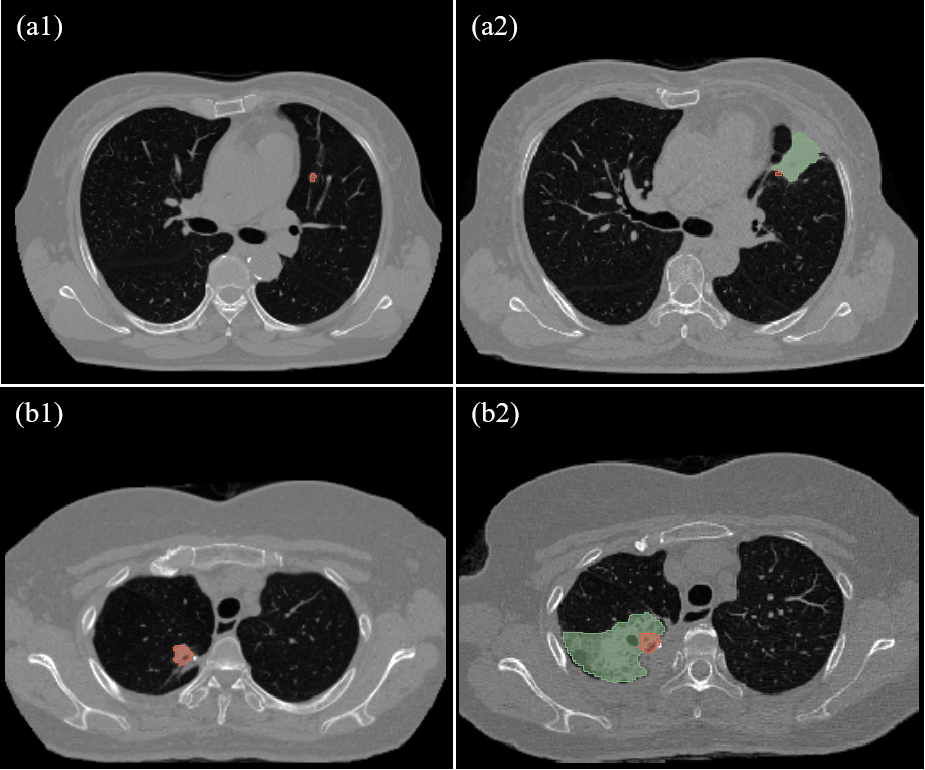}
\caption{Illustration of the two special cases. (a1) and (b1) represent the preoperative CT images of two cases, with the tumor areas indicated in red. (a2) and (b2) represent the corresponding postoperative CT images, where the green areas denote the treatment regions and the red areas represent the registered tumor areas.}
\label{special_case}
\end{figure}

Notably, our experiments identified several outlier data points among cases classified as over-ablation (AES index value of 1) by doctors on immediate postoperative images. Our system predicted an AES index value significantly less than 0, approaching -1, indicating potential under-ablation and possible therapeutic failure. Two such cases are illustrated in Figure \ref{special_case}. Both cases have small tumor areas and large ablation areas, which led doctors to believe the ablation surgery was successful. However, as observed in (a2) and (b2) of Figure \ref{special_case}, the registered tumor areas (red) are not included in the large ablation areas (green), indicating under-ablation. This is also objectively confirmed by medical records, which revealed that one patient was identified as under-ablated during the 1 month follow-up, while another patient was identified as under-ablated during the 6 months follow-up and subsequently underwent another ablation procedure. 

%Figure \ref{data_distribution} illustrates the class distribution between doctors' assessments. The longer-term postoperative ratings paired with medical records can be considered as the ground truth assessment since it was based on all available information and the subsequent results of the therapy. The over-proportion of the over-ablation labels in the left part of the figure shows that, on immediate postoperative images, doctors tend to overestimate the ablation ratings and many cases that should be labeled as under-ablation or average-ablation were miss-labeled as over-ablation. The possible reason is that by just looking at the immediate postoperative image, doctors may not be able to remember the exact location of the preoperative cancer area due to lung deformations, therefore they may tend to assess the treatment majorly based on the absolute ablation area. Table \ref{confusion_doc} shows the confusion matrices between the doctors' subjective scores immediately after surgery and 6 months after. In Table \ref{confusion}, we can find that more than 78\% of cases judged as over-ablation by doctors immediately after surgery were later on rated as under-ablation or average-ablation. This proves that doctors tend to overestimate the ablation ratings when merely judging from the images immediately after surgery.

%This phenomenon suggests that knowing the exact location of the original cancer in postoperative images is vital in preventing misjudgments. Our method provides both image and score to assist the doctors so that they can better understand the relative position between the postoperative ablation area and the preoperative cancer area immediately after surgery. These cases also prove the importance of our registration system which can provide the exact location of cancer even after it disappeared after the ablation. Our respiratory subtraction image could give doctors a solid referencing point when assessing whether the ablation treatment is sufficient and therefore prevent misjudgment.

As shown in Table \ref{spearman}, the Spearman correlation analysis results indicate a robust positive correlation between the AES index and the physicians' assessments (6 months after treatment), which suggests that the AES index can provide valuable support in assessing clinical ablation procedures. Compared with the Spearman correlation coefficient value of 0.443 from the immediate postoperative result, the value of 0.809 indicates that our method shows a stronger correlation with the accurate long-term judgment results. Meanwhile, Table \ref{confusion} shows a confusion matrix between the doctors' subjective scores 6 months after and our AES index (thresholded by $\lambda$). 

%Table \ref{confusion_doc} shows the confusion matrices between the doctors' subjective scores immediately after surgery and 6 months after. In Table \ref{confusion}, we can find that more than 78\% of cases judged as over-ablation by doctors immediately after surgery were later on rated as under-ablation or average-ablation. This proves that doctors tend to overestimate the ablation ratings when merely judging from the images immediately after surgery. 

From table \ref{confusion} we can see that our method can predict the under-ablation cases with full precision, with all the under-ablation cases in the experiment dataset accurately found and rated. As mentioned, under-ablation scenarios can potentially cause dangerous mistreatment or readmission to excessive therapies afterwards. This result shows that our method could preemptively prevent such possibilities caused by doctors' overlook of under-ablation scenarios and protect patients from missing the best treatment opportunity or being subjected to unnecessary second ablation therapy afterward. Compared with Table \ref{confusion_doc} and Table \ref{confusion}, it shows that the result of the AES index is much closer to the doctors' ground-truth assessment (6 months after treatment).

\section{Conclusion and Discussion}

In this study, we proposed the \textit{Respiratory Differencing} method and AES metric to evaluate lung tumor ablation therapy.   In our system, preoperative images are aligned with postoperative images using both coarse rigid and the deedsBCV non-rigid registration method. Subsequently, the registered preoperative images are differentiated from the postoperative images to obtain the \textit{Respiratory Differencing} images. To enhance the clinical assessment of MWA treatment, we introduce a quantitative analysis scheme to evaluate discrepancies between preoperative tumor regions and postoperative treatment coverage. Experimental results using clinical data confirm the effectiveness of both the \textit{Respiratory Differencing} images and our proposed criterion AES index for assessing the ablation area effectiveness during lung tumor ablation surgery.

\section{References}
\bibliographystyle{vancouver}
\bibliography{ref}

\end{document}